\documentclass{article}

\usepackage{spconf,amsmath,graphicx}
\usepackage[backref]{hyperref} 
\usepackage{booktabs} % 导入三线表需要的宏包
\usepackage{longtable}% 导入跨页表格所需宏包
\usepackage{multirow}
\usepackage{graphicx}
\usepackage{float}
\usepackage{amssymb}
\usepackage{color}
\usepackage{wrapfig}

\usepackage{subfigure}
\usepackage{graphicx}
\usepackage{epstopdf}
\usepackage{epsfig}
\usepackage[english]{babel}

\title{Exploiting semantic attributes for Transductive zero-shot learning}

\name{Zhengbo Wang$^{1,2}$, Jian Liang$^{2*}$\thanks{$^*$ corresponding author (liangjian92@gmail.com)}, Zilei Wang$^1$, and Tieniu Tan$^{2,3}$}
\address{
$^1$ University of Science and Technology of China, $^2$ CRIPAC \& MAIS, Institute of Automation,\\ Chinese Academy of Sciences, $^3$ Nanjing University\\}

\begin{document}
% \ninept
%
\maketitle

\begin{abstract}
Zero-shot learning (ZSL) aims to recognize unseen classes by generalizing the relation between visual features and semantic attributes learned from the seen classes.
A recent paradigm called transductive zero-shot learning further leverages unlabeled unseen data during training and has obtained impressive results.
These methods always synthesize unseen features from attributes through a generative adversarial network to mitigate the bias towards seen classes.
However, they neglect the semantic information in the unlabeled unseen data and thus fail to generate high-fidelity attribute-consistent unseen features.
To address this issue, we present a novel transductive ZSL method that produces semantic attributes of the unseen data and imposes them on the generative process.
In particular, we first train an attribute decoder that learns the mapping from visual features to semantic attributes.
Then, from the attribute decoder, we obtain pseudo-attributes of unlabeled data and integrate them into the generative model, which helps capture the detailed differences within unseen classes so as to synthesize more discriminative features.
Experiments on five standard benchmarks show that our method yields state-of-the-art results for zero-shot learning.
\end{abstract}

\begin{keywords}
Zero-shot learning, transductive learning
\end{keywords}

\section{introduction}
\label{intro}
Deep learning has achieved remarkable advances across a wide range of computer vision tasks.
However, these improvements always rely on a fragile assumption under which annotated data is abundant and easy to obtain.
In fact, millions of categories exist in the real world, and new concepts emerge every day. 
Therefore, collecting annotated data for all categories is unrealistic.

Zero-shot learning (ZSL)~\cite{lampert2009learning} resolves the issue by striving to recognize unseen objects for which no labeled data is available during training.
To accomplish this, ZSL approaches leverage extra semantic information to bridge the gap between seen and unseen classes, which could be semantic attributes~\cite{farhadi2009describing}, word embeddings~\cite{mikolov2013efficient}, or language descriptions~\cite{reed2016learning} of classes.
Hence, they can recognize the unseen classes by generalizing the visual-semantic relation learned from seen classes.
Typically, ZSL methods could be divided into two main categories: inductive ZSL where only labeled seen data is available during training, and transductive ZSL where both labeled seen data and unlabeled unseen data are available. 
Inductive ZSL methods have been found to yield biased predictions towards seen classes~\cite{xian2018feature, han2021contrastive, Huynh-DAZLE:CVPR20}, which is unsatisfactory for real-world applications.
Recent transductive ZSL methods~\cite{narayan2020latent, xian2019f, wu2020self} mitigate the bias by incorporating the unlabeled data into the popular generative modeling framework, generative adversarial network (GAN)~\cite{goodfellow2014generative}. 
Generally, there exists a generator, a conditional discriminator, and an unconditional discriminator, creating a Y-shape framework.
The shared generator and two branched discriminators form a conditional and an unconditional GAN for seen and unseen classes, respectively.
Taking seen images and matched attributes as input, the conditional discriminator discriminates whether the images come from the conditional distribution of the corresponding attributes.
Contrarily, with unlabeled unseen images, the unconditional discriminator simply distinguishes whether the images belong to unseen classes distribution, serving as regularization to alleviate the bias problem.
Since they neglect to model the conditional distribution of unseen classes like the conditional GAN, the generated features of different unseen classes may congregate.

To deal with the problem, we propose a pseudo-attribute-based method, which imposes class-wise constraints on the unseen discriminator by exploiting semantic attributes.
First, we train a Y-shape generative model in company with an auxiliary attribute decoder, which learns the mapping from visual features to semantic attributes.
With the pretrained attribute decoder, we can produce reliable pseudo-attributes to represent the class information of the unseen data.
Taking the unseen data and the corresponding pseudo-attribute as input, the unseen discriminator can discriminate whether the data is real and consistent with the attribute, and therefore modeling the conditional distribution.
% Hence, our model synthesizes more compact and discriminative features.
Experiments on standard benchmarks validate the superior performance of our method over the state-of-the-art methods.

\section{Proposed Method}
\label{method}
\begin{figure*}[htbp]
    \centering
    \includegraphics[width=1.0\textwidth]{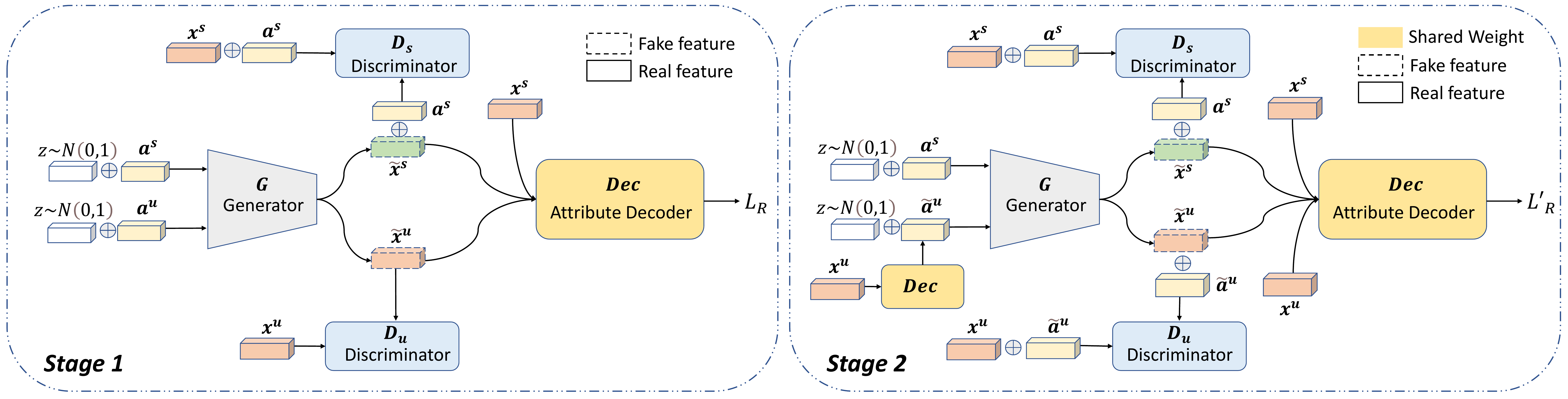}
    \caption{
    \textbf{The overall architecture of our method.} 
    It consists of one generator $G$ and two discriminators, $D_s$ and $D_u$, for seen and unseen classes, respectively. 
    Moreover, an auxiliary attribute decoder $Dec$ is introduced to learn to generate semantically consistent features. 
    (1) In stage 1, we train a generative model with the unconditional unseen discriminator $D_u$ in the absence of matched semantic attributes of unseen data. 
    (2) In stage 2, we employ the pretrained attribute decoder $Dec$ to exploit the class-wise information by generating pseudo-attributes for unseen data, which is then concatenated into the generative process.
    Finally, we retrain the modified generative model and obtain the generator for the ZSL classification.
    }
    \label{fig:architecture}
    \vspace{-10pt}
\end{figure*}

% We begin by formulating transductive ZSL. 
Assuming that we have two disjoint sets of classes: seen classes $\mathcal{Y}_s$ and unseen classes $\mathcal{Y}_u$, where $\mathcal{Y}_s\cap\mathcal{Y}_u = \emptyset$.
For the seen classes, we have labeled data: $\mathcal{S}$ = $\{(x_i^s, y_i^s, a_i^s)\}_{i=1}^{N_s}$, where $x_i^s\in\mathcal{X}$ and $a_i^s\in\mathcal{A}_s $ denote its $d$-dimension visual feature and $k$-dimension semantic attribute for the $i$-th image with label $y_i^s\in\mathcal{Y}_s$.
And for the unseen classes, an extra unlabeled unseen data is available: $\mathcal{U}$ = $\{x_i^u\}_{i=1}^{N_u}$, where $x_i^u\in\mathcal{X}$ is $i$-th unseen visual feature. Moreover, we have a set of unseen attributes $\mathcal{A}_u$.
We denote $\mathcal{A} = \mathcal{A}_s\cup\mathcal{A}_u$ as the set of attributes of all classes. 

\subsection{Pseudo-Attribute-Based Method}
\label{subsec:pseudo}
The overall architecture of our method is shown in Figure~\ref{fig:architecture}.
We base our approach on a generative adversarial network, which has shown strong performance in recent ZSL methods~\cite{narayan2020latent, xian2018feature, xian2019f, wu2020self, Sariyildiz_2019_CVPR}. 
As illustrated in Figure~\ref{fig:architecture}, our method contains two stages. 
In stage 1, an attribute decoder is trained with a generative model, learning to generate semantic attributes from visual features.
Then, in stage 2, we produce pseudo-attributes for the unlabeled unseen data and integrate them into the generative process to exploit the semantic information in unseen classes, which helps synthesize semantically consistent unseen features. 
Finally, we utilize the generator to synthesize unseen features for ZSL classification.

\textbf{Stage 1}. 
As shown in Figure~\ref{fig:architecture}, we train a three-branch GAN~\cite{wgan}, which consists of a generator, an attribute decoder, and two discriminators. 
The generator $G(z, a)$ synthesizes a feature from random noise and an attribute. 
The conditional seen discriminator $D_s(x, a)$ takes a seen feature and its corresponding attribute as input and outputs a real value indicating the degree of realness or fakeness. 
The unconditional unseen discriminator $D_u(x)$ takes an unseen feature as input and outputs its realness/fakeness score for unseen classes. 
For seen class data $(x^s, y^s, a^s)\in\mathcal{S}$, since we have matched feature and attribute, the generator $G$ and the seen discriminator $D_s$ construct a conditional GAN, and it is optimized by the adversarial loss:
\begin{equation}
\begin{split}
    L_S & = \mathbb{E}[D_s(x^s, a^s)] - \mathbb{E}[D_s(\Tilde{x}^s, a^s)] \\
        & - \lambda\mathbb{E}[(\|\nabla D_s(\hat{x}^s, a^s)\|_2 - 1)^2],
\end{split}
\end{equation}
where $\Tilde{x}^s = G(z, a^s), z\sim\mathcal{N}(0, 1)$ is the synthesized feature, $\lambda$ is the penalty coefficient, and $\hat{x}^s = \alpha x^s + (1 - \alpha)\Tilde{x}^s, \alpha\sim U(0,1)$. For the unseen feature $x^u\in\mathcal{U}$,  we randomly sample an unseen attribute $a^u\in\mathcal{A}_u$. Since the feature and attribute are mismatched, the generator $G$ and the unseen discriminator $D_u$ construct an unconditional GAN, and it is optimized by:
\begin{equation}
\begin{split}
    L_U & = \mathbb{E}[D_u(x^u)] - \mathbb{E}[D_u(\Tilde{x}^u)] \\
        & - \lambda\mathbb{E}[(\|\nabla D_u(\hat{x}^u)\|_2 - 1)^2], 
\end{split}
\end{equation} 
where $\Tilde{x}^u = G(z, a^u), z\sim\mathcal{N}(0, 1)$, $\lambda$ is the penalty coefficient, and $\hat{x}^u = \beta x^u + (1 - \beta)\Tilde{x}^u, \beta\sim U(0, 1)$. In order to generate semantically consistent features and generate pseudo-attributes for the next stage, we add an auxiliary attribute decoder $Dec: \mathcal{X}\to\mathcal{A}$. And the cycle-consistency of the attributes is achieved by using $\ell_1$ reconstruction loss:
\begin{equation}
\begin{split}
    L_R & = \mathbb{E}[\|Dec(x^s) - a^s\|_1] + \mathbb{E}[\|Dec(\Tilde{x}^s) - a^s\|_1] \\
        & + \mathbb{E}[\|Dec(\Tilde{x}^u) - a^u\|_1].
\end{split}
\end{equation}
Hence, the overall model is optimized by combining adversarial losses and the reconstruction loss as follows:
\begin{equation}
    \min_{D_s, D_u, Dec}\max_{G} L_S + L_U + w L_R,
\end{equation}
where $w$ is a hyper-parameter for weighting the reconstruction loss.

\textbf{Stage 2}. 
From the well-trained attribute decoder, we can obtain pseudo-attributes of the unlabeled data and utilize them as the conditional information of the unseen discriminator, i.e., we transform the unconditional discriminator $D_u(x)$ to a conditional one $D_u(x, a)$ to exploit the class-wise information of the unseen data. 
After the transformation, we reinitialize the generator $G$ and the two discriminators, $D_s$ and $D_u$, to make the generator and the discriminators compatible.
Only the attribute decoder $Dec$ maintains the weights trained in stage 1. 
To optimize the discriminator $D_u(x, a)$, we first obtain the pseudo-attribute $\Tilde{a}^u = Dec(x^u)$ and consider it to be the corresponding attribute of the unseen data.
And we concatenate and input them into $D_u(x, a)$ as~\cite{hu2021adversarial}.
% Similar to~\cite{hu2021adversarial}, we first obtain the pseudo-attribute $\Tilde{a}^u = Dec(x^u)$ of unseen data and concatenate with the unseen data and then input them to the unseen discriminator $D_u(x, a)$.
% Similar to~\cite{hu2021adversarial}, we concatenate the unseen data and the pseudo-attribute $\Tilde{a}^u = Dec(x^u)$ 
Thus the adversarial loss is modified as follows:
\begin{equation}
\begin{split}
    L'_U & = \mathbb{E}[D_u(x^u, \Tilde{a}^u)] - \mathbb{E}[D_u(\Tilde{x}^u, \Tilde{a}^u)] \\
        & - \lambda\mathbb{E}[(\|\nabla D_u(\hat{x}^u, \Tilde{a}^u)\|_2 - 1)^2],
\end{split}
\end{equation}
where $\Tilde{x}^u = G(z, \Tilde{a}^u), z\sim\mathcal{N}(0, 1)$, and $\hat{x}^u = \gamma x^u + (1 - \gamma)\Tilde{x}^u, \gamma\sim U(0, 1)$. Furthermore, we modify the reconstruction loss by replacing $a^u$ with $\Tilde{a}^u$. Thus the attribute decoder is fine-tuned by:
\begin{equation}
\begin{split}
    L'_R & = \mathbb{E}[\|Dec(x^s) - a^s\|_1] + \mathbb{E}[\|Dec(\Tilde{x}^s) - a^s\|_1] \\
        & + \mathbb{E}[\|Dec(\Tilde{x}^u) - \Tilde{a}^u\|_1].
\end{split}
\end{equation}
To this end, our full objective in stage 2 becomes:
\begin{equation}
    \min_{D_s, D_u, Dec}\max_{G}  L_S + L'_U + w' L'_R,
\end{equation}
and $w'$ is the loss coefficient.

\subsection{Zero-Shot Classification}
\label{subsec:classification}
Once the generator is trained, we leverage it to synthesize unseen-class features $\mathcal{\Tilde{U}}$ from corresponding unseen attributes in order to train the ZSL classifier.
Typically, there are two kinds of test protocol in ZSL: conventional ZSL and generalized ZSL. 
For conventional ZSL, test samples only come from unseen categories.
Thus, we use $\mathcal{\Tilde{U}}$ to train a classifier by minimizing the cross-entropy loss: $\min_\theta-\frac{1}{|\mathcal{X}|}\sum_{(x,y)\sim(\mathcal{X}, \mathcal{Y})}\log P(y|x;\theta),$
where $P(y|x;\theta) = \frac{\exp(\theta_y^Tx)}{\sum_{j=1}^{|\mathcal{Y}|}\exp(\theta_j^Tx)}$ is classification probability and $\theta$ denotes classifier parameters. 
For generalized ZSL, test samples come from both seen and unseen classes.
To deal with the bias problem, we adopt the cascaded classifier mentioned in~\cite{liu2022hardboost, bo2021hardness}.
We first train a binary classifier with $\mathcal{U}$ and $\mathcal{S}$, which coarsely classify samples into seen or unseen classes.
Then, we input the `seen-class' data to the seen-class classifier and the `unseen-class' data to the unseen-class classifier.

\begin{table}[htbp]
  \centering
  \caption{
  \textbf{Conventional zero-shot learning performance.} 
  Both inductive (IN) and transductive (TR) results are shown in terms of average per-class top-1 accuracy.
  \textcolor{red}{Red} and \textcolor{blue}{blue} denote the highest and second highest results.
  }
  \scalebox{0.83}{
    \begin{tabular}{cl|ccccc}
    \toprule
    \multicolumn{2}{c|}{} & \textbf{CUB} & \textbf{AWA1} & \textbf{AWA2} & \textbf{SUN} & \textbf{FLO} \\
    \midrule
    \multirow{6}[1]{*}{IN} 
          & DeViSE~\cite{frome2013devise} & 52.0  & 54.2  & 59.7  & 56.5  & 45.9 \\
          & DAZLE~\cite{Huynh-DAZLE:CVPR20} & 65.9  & -     & -     & -     & - \\
          & CE-GZSL~\cite{han2021contrastive} & 77.5  & 71.0  & 70.4  & 63.3  & 70.6 \\
          & LisGAN~\cite{li2019leveraging} & 58.8  & 70.6  & -     & 61.7  & 69.6 \\
          & IZF-Softmax~\cite{shen2020invertible} & -     & -     & -     & -     & - \\
          & f-CLSGAN~\cite{xian2018feature} & 57.3  & 68.2  & -     & 60.8  & 67.2 \\
    \midrule
    \multirow{7}[2]{*}{TR} 
          & GXE~\cite{li2019rethinking}   & 61.3  & 89.8  & 83.2  & 63.5  & - \\
          & GMN~\cite{Sariyildiz_2019_CVPR}   & 64.6  & 82.5  & -     & 64.3  & - \\
          & f-VAEGAN-D2~\cite{xian2019f} & 71.1  & -     & 89.8  & 70.1  & 89.1 \\
          & SGDN~\cite{wu2020self}  & 74.9  & \textcolor[rgb]{ 0,  0,  1}{92.3}  & 93.4  & 68.4  & 81.8 \\
          & STHS-GAN~\cite{bo2021hardness} & \textcolor[rgb]{ 0,  0,  1}{77.4}  & -     & \textcolor[rgb]{ 1,  0,  0}{94.9} & 67.5  & - \\
          & TF-VAEGAN~\cite{narayan2020latent} & 74.7  & -     & 92.1  & \textcolor[rgb]{ 0,  0,  1}{70.9}  & \textcolor[rgb]{ 0,  0,  1}{92.6} \\
          & \textbf{ours} & \textcolor[rgb]{ 1,  0,  0}{79.7} & \textcolor[rgb]{ 1,  0,  0}{94.2} & \textcolor[rgb]{ 0,  0,  1}{94.2}  & \textcolor[rgb]{ 1,  0,  0}{73.8} & \textcolor[rgb]{ 1,  0,  0}{94.9} \\
    \bottomrule
    \end{tabular}%
    }
  \label{tab:zsl}%
  \vspace{-10pt}
\end{table}%

\section{Experiments}
\textbf{Datasets.} We evaluate our method on five standard zero-shot object recognition datasets: CUB~\cite{wah2011caltech}, FLO~\cite{nilsback2008automated}, SUN~\cite{patterson2012sun}, AWA1~\cite{lampert2009learning}, and AWA2~\cite{XLSA18}, containing 200, 102, 717, 50, and 50 categories, respectively. 
We follow the same split setting, evaluation metric, and semantic embedding mentioned in~\cite{XLSA18} for a fair comparison.

\noindent 
\textbf{Evaluation Metric.}
In conventional ZSL, where test images only come from unseen classes, we measure it by the average per-class top-1 accuracy.
In generalized ZSL, test images come from either seen or unseen classes. 
Thus it is measured by the average per-class top-1 accuracy on seen and unseen classes, denoted as $S$ and $U$, respectively. 
Furthermore, to measure the comprehensive performance, we use the harmonic mean as the final metrics: $H = \frac{2*U*S}{U+S}$. 

\noindent 
\textbf{Implementation Details.} 
Following~\cite{narayan2020latent, xian2019f}, we adopt VAEGAN as our generative model.
The generator $G$, discriminators $D_s, D_u$, and attribute decoder $Dec$ are implemented as two-layer fully-connected networks with 4,096 hidden dimensions. 
The classifiers at the test stage input the transformed features mentioned in~\cite{narayan2020latent}.
The dimension of random noise $z$ is equal to the attribute dimension ($\mathbb{R}^{d_z} = \mathbb{R}^{d_a}$). 
We use Adam optimizer with $10^{-4}$ learning rate. 
We set hyper-parameters $\lambda = 5$, $w = 0.1$, and $w' = 0.1$ following~\cite{narayan2020latent}.

\subsection{State-of-the-art Comparison}

% Table generated by Excel2LaTeX from sheet 'Sheet1'
\setlength{\tabcolsep}{7.5pt}
\begin{table*}[htbp]
  \centering
  \caption{
  \textbf{Generalized zero-shot learning performance.} 
  Both inductive (IN) and transductive (TR) results are shown.
  U = unseen acc, S = seen acc, \textbf{H} = harmonic mean of seen and unseen accs.
  }
  \resizebox{1.0\textwidth}{!}{
    \begin{tabular}{cl|ccc|ccc|ccc|ccc|ccl}
    \toprule
    \multicolumn{2}{c|}{\multirow{2}[3]{*}{Method}} & \multicolumn{3}{c|}{\textbf{CUB}} & \multicolumn{3}{c|}{\textbf{AWA1}} & \multicolumn{3}{c|}{\textbf{AWA2}} & \multicolumn{3}{c|}{\textbf{SUN}} & \multicolumn{3}{c}{\textbf{FLO}} \\
\cmidrule{3-17}    \multicolumn{2}{c|}{} & U     & S     & \textbf{H} & U     & S     & \textbf{H} & U     & S     & \textbf{H} & U     & S     & \textbf{H} & U     & S     & \multicolumn{1}{c}{\textbf{H}} \\
    \midrule
    \multirow{6}[1]{*}{IN} 
          & DeViSE~\cite{frome2013devise} & 23.8  & 53.0  & 32.8  & 13.4  & 68.7  & 22.4  & 17.1  & 74.7  & 27.8  & 16.9  & 27.4  & 20.9  & 9.9   & 44.2  & 16.2 \\
          & DAZLE~\cite{Huynh-DAZLE:CVPR20} & 56.7  & 59.6  & 58.1  & -     & -     & -     & 60.3  & 75.7  & 67.1  & 52.3  & 24.3  & 33.2  & -     & -     & - \\
          & CE-GZSL~\cite{han2021contrastive} & 63.9  & 66.8  & 65.3  & 65.3  & 73.4  & 69.1  & 63.1  & 78.6  & 70.0  & 48.8  & 38.6  & 43.1  & 69.0  & 78.7  & 73.5 \\
          & LisGAN~\cite{li2019leveraging} & 46.5  & 57.9  & 51.6  & 52.6  & 76.3  & 62.3  & -     & -     & -     & 42.9  & 37.8  & 40.2  & 57.7  & 83.8  & 68.3 \\
          & IZF-Softmax~\cite{shen2020invertible} & 52.7  & 68.0  & 59.4  & 61.3  & 80.5  & 69.6  & 60.6  & 77.5  & 68.0  & 52.7  & 57.0  & 54.8  & -     & -     & - \\
          & f-CLSGAN~\cite{xian2018feature} & 43.7  & 57.7  & 49.7  & 57.9  & 61.4  & 59.6  & 52.1  & 68.9  & 59.4  & 42.6  & 36.6  & 39.4  & 59.0  & 73.8  & 65.6 \\
    \midrule
    \multirow{7}[2]{*}{TR} 
          & GXE~\cite{li2019rethinking} & 57.0  & 68.7  & 62.3  & \textcolor[rgb]{ 0,  0,  1}{87.7}  & \textcolor[rgb]{ 0,  0,  1}{89.0}  & \textcolor[rgb]{ 0,  0,  1}{88.4}  & 80.2  & 90.0  & 84.8  & 45.4  & \textcolor[rgb]{ 1,  0,  0}{58.1} & 51.0  & -     & -     & - \\
          & GMN~\cite{Sariyildiz_2019_CVPR} & 60.2  & 70.6  & 65.0  & 70.8  & 79.2  & 74.8  & -     & -     & -     & 57.1  & 40.7  & 47.5  & -     & -     & - \\
          & f-VAEGAN-D2~\cite{xian2019f} & 61.4  & 65.1  & 63.2  & -     & -     & -     & 84.8  & 88.6  & 86.7  & 60.6  & 41.9  & 49.6  & 78.7  & 87.2  & 82.7 \\
          & SGDN~\cite{wu2020self} & 69.9  & 70.2  & 70.1  & 87.3  & 88.1  & 87.7  & 88.8  & 89.3  & 89.1  & 62.0  & 46.0  & 52.8  & 78.3  & 91.4  & 84.4 \\
          & STHS-GAN~\cite{bo2021hardness} & \textcolor[rgb]{ 0,  0,  1}{77.4}  & \textcolor[rgb]{ 1,  0,  0}{74.5} & \textcolor[rgb]{ 0,  0,  1}{72.8}  & -     & -     & -     & \textcolor[rgb]{ 1,  0,  0}{94.9} & \textcolor[rgb]{ 1,  0,  0}{92.3} & \textcolor[rgb]{ 1,  0,  0}{93.6} & \textcolor[rgb]{ 0,  0,  1}{67.5}  & 44.8  & \textcolor[rgb]{ 0,  0,  1}{53.9}  & -     & -     & - \\
          & TF-VAEGAN~\cite{narayan2020latent} & 69.9  & 72.1  & 71.0  & -     & -     & -     & 87.3  & 89.6  & 88.4  & 62.4  & 47.1  & 53.7  & \textcolor[rgb]{ 0,  0,  1}{91.8}  & \textcolor[rgb]{ 0,  0,  1}{93.2}  & \textcolor[rgb]{ 0,  0,  1}{92.5} \\
          & \textbf{ours} & \textcolor[rgb]{ 1,  0,  0}{79.3} & \textcolor[rgb]{ 0,  0,  1}{73.2}  & \textcolor[rgb]{ 1,  0,  0}{76.1} & \textcolor[rgb]{ 1,  0,  0}{94.1} & \textcolor[rgb]{ 1,  0,  0}{90.0} & \textcolor[rgb]{ 1,  0,  0}{92.0} & \textcolor[rgb]{ 0,  0,  1}{93.6}  & \textcolor[rgb]{ 0,  0,  1}{91.7}  & \textcolor[rgb]{ 0,  0,  1}{92.6}  & \textcolor[rgb]{ 1,  0,  0}{72.4} & \textcolor[rgb]{ 0,  0,  1}{48.9}  & \textcolor[rgb]{ 1,  0,  0}{58.4} & \textcolor[rgb]{ 1,  0,  0}{94.6} & \textcolor[rgb]{ 1,  0,  0}{93.9} & \textcolor[rgb]{ 1,  0,  0}{94.3} \\
    \bottomrule
    \end{tabular}%
    }
    \vspace{-10pt}
  \label{tab:gzsl}%
\end{table*}%

We select several advanced inductive ZSL and transductive ZSL methods for comparison.
(1) For inductive ZSL, we choose six competitive methods, including DeViSE~\cite{frome2013devise}, DAZLE~\cite{Huynh-DAZLE:CVPR20}, CE-GZSL~\cite{han2021contrastive}, LisGAN~\cite{li2019leveraging}, IZF-Softmax~\cite{shen2020invertible}, and f-CLSGAN~\cite{xian2018feature}. 
(2) For transductive ZSL, we choose six state-of-the-art methods, including GXE~\cite{li2019rethinking}, GMN~\cite{Sariyildiz_2019_CVPR}, f-VAEGAN-D2~\cite{xian2019f}, SGDN~\cite{wu2020self}, STHS-GAN~\cite{bo2021hardness}, and TF-VAEGAN~\cite{narayan2020latent}. Our method falls into the transductive type.

\begin{figure}[htbp]
    \centering
    \subfigure[TF-VAEGAN synthesized features]{
        \begin{minipage}[t]{0.5\linewidth}
            \includegraphics[width=\linewidth]{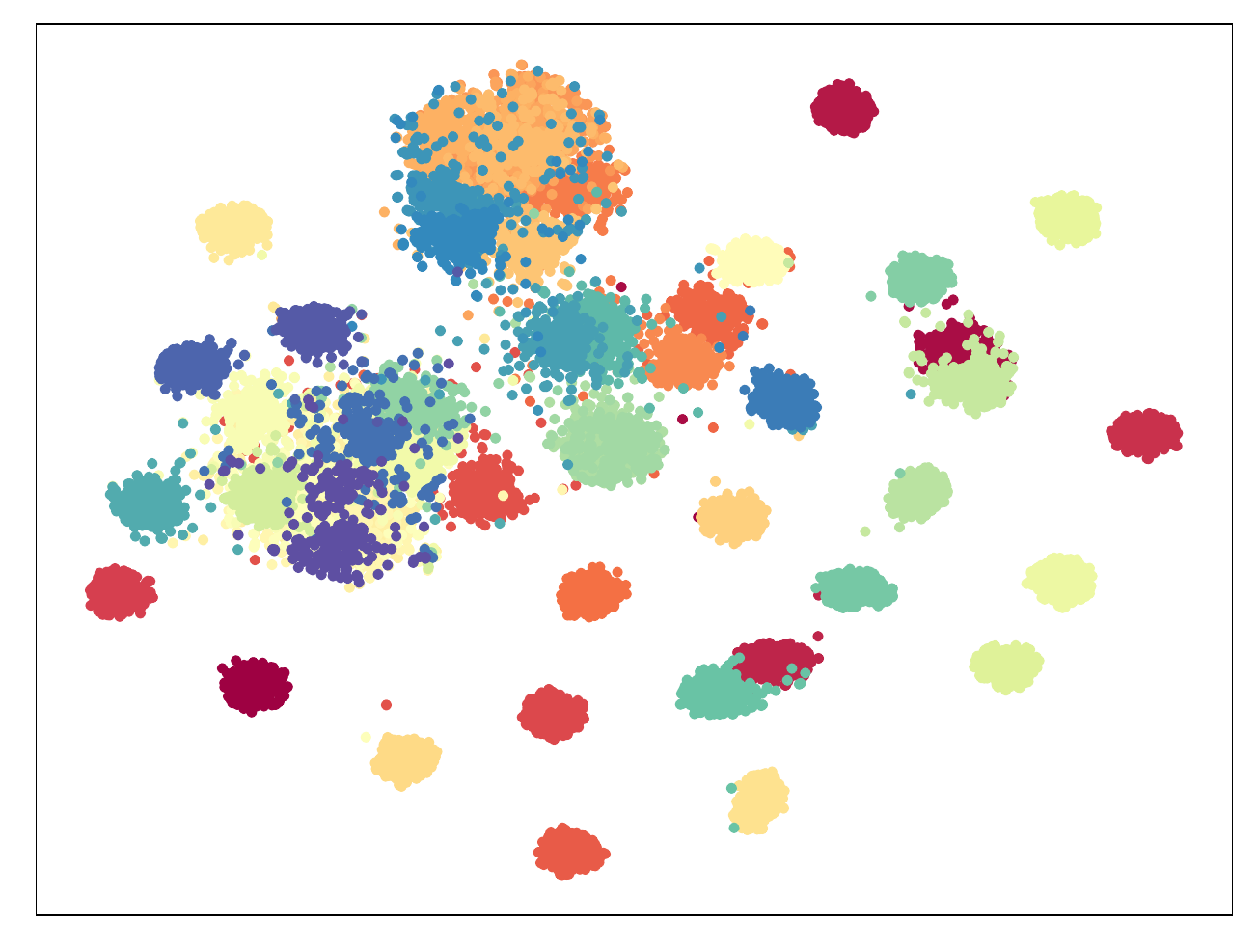} 
            % \caption{caption1} 
            \label{tsne_f1}
        \end{minipage}%
    }%
    \subfigure[Our method synthesized features]{
        \begin{minipage}[t]{0.5\linewidth}
            \includegraphics[width=\linewidth]{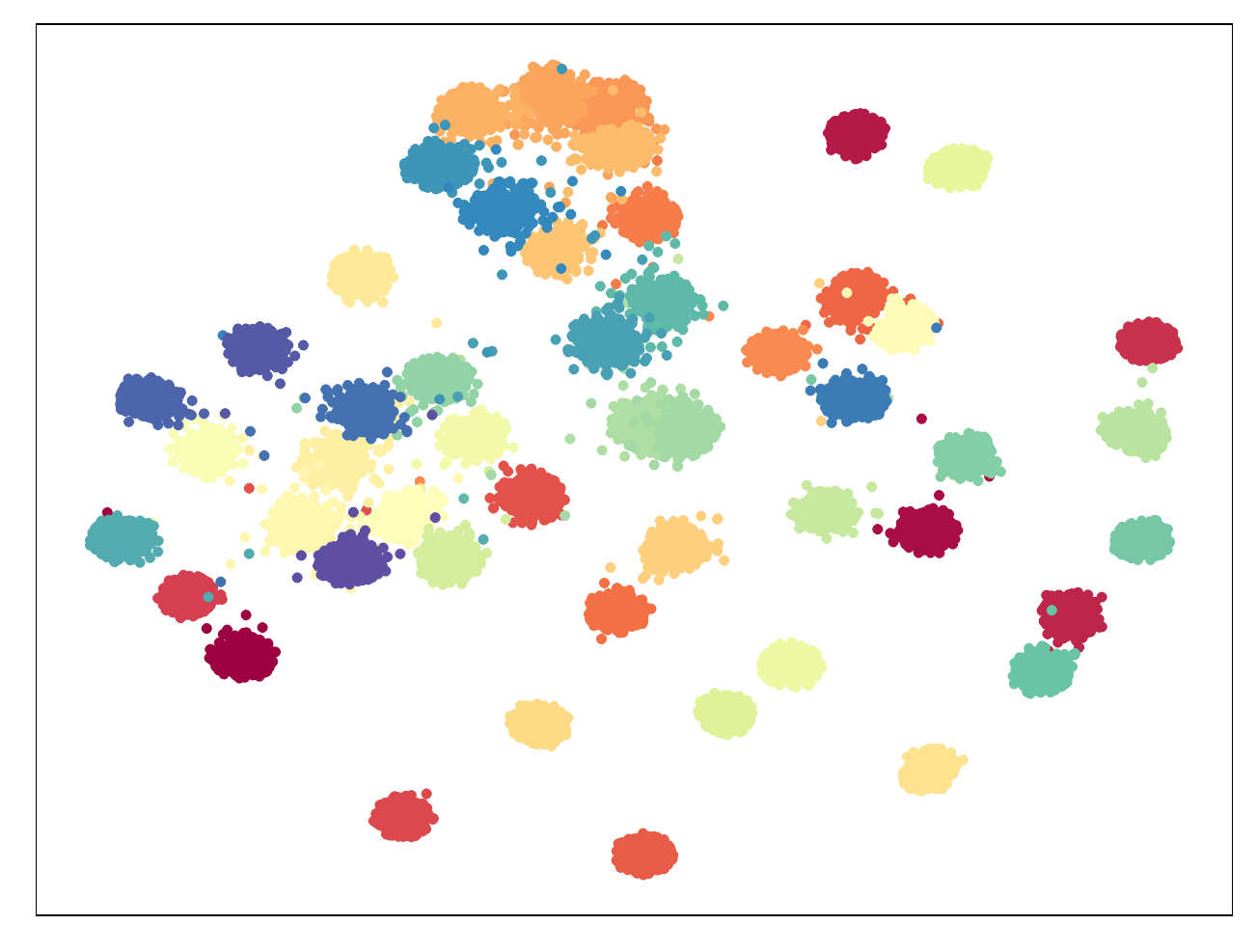}
            % \caption{caption2}
            \label{tsne_f2}
        \end{minipage}
    }%
    \caption{
    The t-SNE visualization of unseen features synthesized by (a) TF-VAEGAN~\cite{narayan2020latent} and (b) our method on CUB.
    }
    \label{fig:tsne}
    \vspace{-5pt}
\end{figure}

\noindent
\textbf{Results of conventional zero-shot learning.}
As shown in Table~\ref{tab:zsl}, our method achieves state-of-the-art results on five ZSL datasets in competition with both inductive methods and transductive methods. 
Our method acquires excellent performance and establishes new benchmarks on four datasets. 
Especially on CUB, SUN, and FLO, the improvements are more than 2\%. 
The reason is that they are very fine-grained datasets.
In this case, features in different unseen classes are close to each other.
Therefore, with the aid of pseudo-attributes, we synthesize more discriminative features and generate more clear boundaries for unseen classes, which improves ZSL performance.
Moreover, our method consistently outperforms SGDN~\cite{wu2020self}, f-VAEGAN-D2~\cite{xian2019f}, and TF-VAEGAN~\cite{narayan2020latent} on all datasets, because they all neglect the semantic information in unseen classes.

\noindent
\textbf{Results of generalized zero-shot learning.}
We also conduct experiments on generalized ZSL. As shown in Table~\ref{tab:gzsl}, our method achieves either the best or second best results among most of the results of seen accuracy, unseen accuracy, and the harmonic mean. Our method obtains 3.3\%, 3.6\%, 3.6\%, and 1.8\% improvement in harmonic mean for CUB, AWA1, SUN, and FLO datasets, respectively. We attribute this gain to the pseudo-attribute-based method to make clear decision boundaries for different unseen classes.

\subsection{Ablation Study}
\noindent
\textbf{t-SNE Visualization.} 
To further verify our method, we conduct t-SNE~\cite{van2008visualizing} visualization for synthesized unseen features of our method and TF-VAEGAN~\cite{narayan2020latent} on CUB. 
As shown in Figure~\ref{fig:tsne}, compared to TF-VAEGAN~\cite{narayan2020latent}, the synthesized features of our method are more compact and more distinct, especially in the top-left corner of the figure.
\par
% Table generated by Excel2LaTeX from sheet 'Sheet1'
\begin{table}[htbp]
  \centering
  \caption{Top-1 ZSL accuracy of different pretrained epochs at stage 1 on the CUB dataset.}
  \scalebox{1.0}{
    \begin{tabular}{cc|cccccc}
    \toprule
    \multicolumn{2}{c|}{epochs} & 0     & 100   & 200   & 300   & 400   & 500 \\
    \midrule
    \multicolumn{2}{c|}{acc} & 74.5  & 77.9  & 78.4  & 79.2  & 79.3  & 79.7 \\
    \bottomrule
    \end{tabular}%
    }
    % \vspace{-5pt}
  \label{tab:histogram}%
\end{table}%

\noindent
\textbf{Importance of stage 1.}
We take experiments to verify the importance of pretraining at stage 1.
Table~\ref{tab:histogram} shows the conventional ZSL results of our method of CUB dataset with different training epochs in stage 1. 
Table~\ref{tab:histogram} demonstrates that 
(1) without stage 1 (epoch=0), our method performs even worse than TF-VAEGAN~\cite{narayan2020latent},
and (2) including stage 1 in training has a positive effect, and the more epochs we pretrain at stage 1, the better results we finally achieve.

\section{conclusion}
This paper proposes a pseudo-attribute-based method for transductive ZSL.
It exploits semantic information of unseen classes by generating and incorporating the pseudo-attributes into the generative process, which helps synthesize discriminative unseen features.
We conduct extensive experiments to show the superiority of our method over prior methods.
% We conduct extensive experiments in various ZSL settings to show the superiority of our method.

\section{Acknowledgements}
This work was in part funded by National Natural Science Foundation of China (Grant No. 62276256), Beijing Nova Program under Grant Z211100002121108 and CAAI-Huawei MindSpore Open Fund.
Our work is partially based on the MindSpore framework.

% To start a new column (but not a new page) and help balance the last-page
% column length use \vfill\pagebreak.
% -------------------------------------------------------------------------
%\vfill
%\pagebreak

\vfill\pagebreak

% References should be produced using the bibtex program from suitable
% BiBTeX files (here: strings, refs, manuals). The IEEEbib.bst bibliography
% style file from IEEE produces unsorted bibliography list.
% -------------------------------------------------------------------------
\bibliographystyle{IEEEbib}
\bibliography{refs}

\end{document}